\documentclass{esannV2}
\usepackage[dvips]{graphicx}
\usepackage[latin1]{inputenc}
\usepackage{amssymb,amsmath,array}

\usepackage{booktabs}
\graphicspath{ {./figs/} }
\usepackage{amsmath}
\usepackage{tabularx}
\DeclareMathOperator{\atantwo}{atan2}
\usepackage[skip=8pt]{caption}
\usepackage{url}

\begin{document}
\title{Fourier-based Video Prediction through Relational Object Motion}

\author{Malte Mosbach and Sven Behnke

%
\vspace{.3cm}\\
%
University of Bonn, Computer Science Institute VI, Autonomous Intelligent Systems \\
Friedrich-Hirzebruch-Allee 5, 53115 Bonn, Germany \\
\{mosbach, behnke\}@ais.uni-bonn.de
%
}

\maketitle

\begin{abstract}
The ability to predict future outcomes conditioned on observed video frames is crucial for intelligent decision-making in autonomous systems. Recently, deep recurrent architectures have been applied to the task of video prediction. However, this often results in blurry predictions and requires tedious training on large datasets. Here, we explore a different approach by (1) using frequency-domain approaches for video prediction and (2) explicitly inferring object-motion relationships in the observed scene. The resulting predictions are consistent with the observed dynamics in a scene and do not suffer from blur. 
\end{abstract}

\section{Introduction}
Anticipating future events is a key capability for intelligent decision-making systems. Future video frame prediction requires scene understanding and has recently received much attention. The latter task is predominantly solved by deep recurrent architectures that encode frames into a hidden representation and model the temporal evolution in latent space before synthesizing a predicted image. Training such architectures to predict images based on the mean squared error loss in the image space results in blurry predictions. In contrast to this, we explore the approach of modeling time series of images by explicitly predicting the transformations between images. Instead of using learned filters to extract transformations, we estimate phase differences between the frequency-domain representations of consecutive frames. First, to model nonlinear motion, we learn transformations of transformations in a hierarchy. We add a small recurrent neural network, to choose between basic motion patterns. Second, we study the hierarchical relationships of objects in a scene. Our goal is to capture the compositional structure underlying dynamic systems by reducing complex motion patterns to simpler primitive motions and the relationships between objects. To this end, we study synthetic sequences of interacting objects created by a hierarchical probabilistic model.

\begin{figure}[ht!]
\centering
\includegraphics[width=1.0\textwidth]{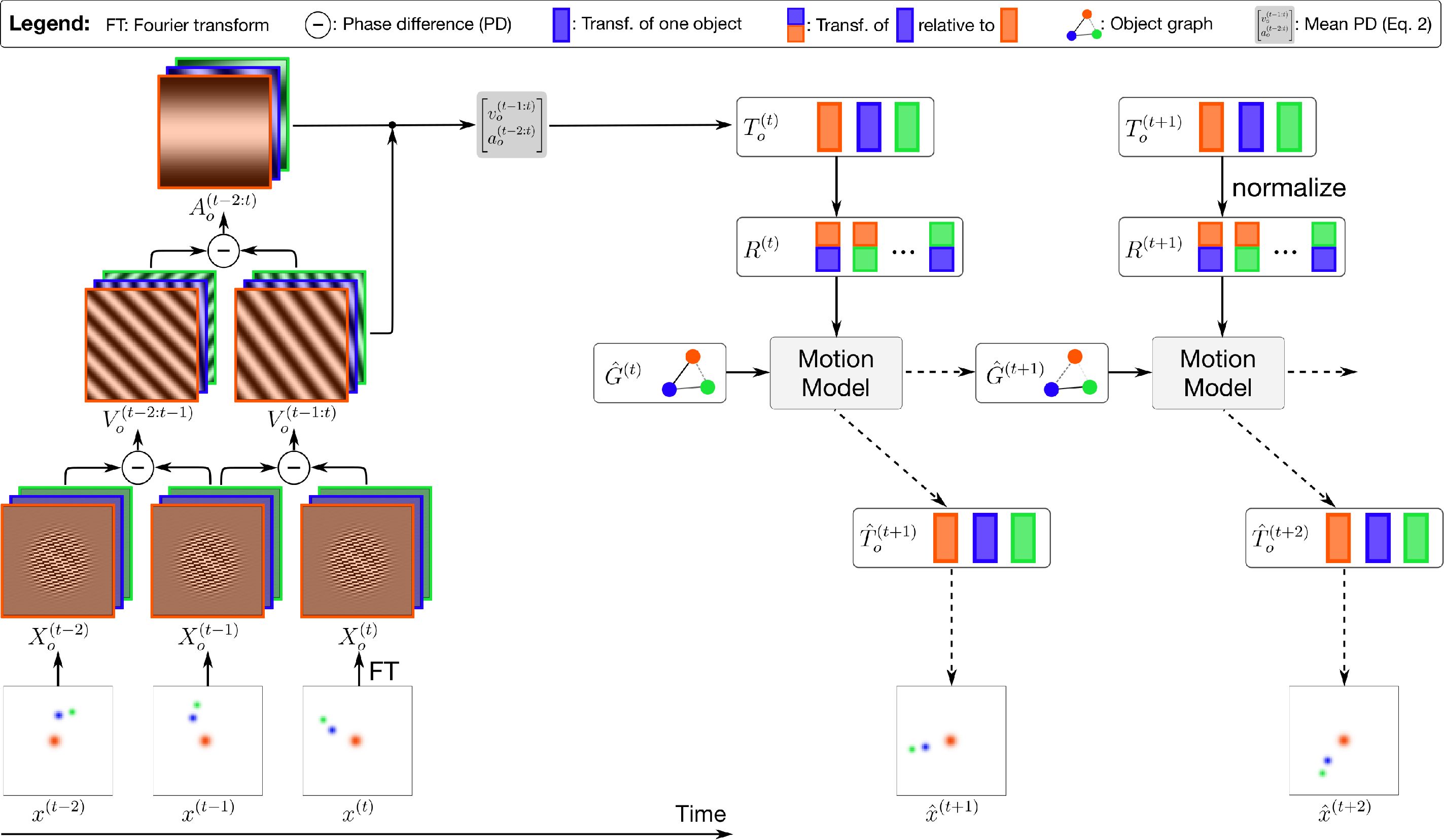}
\caption{Left: Extraction of object transformations in frequency space. Right: Framework for prediction. Observed transformations \(T_{o}^{(t)}\) are normalized by applying the inverse transformations of a parent object and are input at each time step \(t\) into the motion model, which outputs the predicted transformation \(\hat{T}_{o}^{(t+1)}\) for each object \(o\) and an estimate of the object graph \(\hat{G}^{(t)}\). \(\hat{T}_{o}^{(t+1)}=[V^{(t-2:t-1)}, V^{(t-1:t)}, A^{(t-2:t)}]\), where \( [\cdot, \cdot]\) denotes concatenation. \(R^{(t)}\) represents the set of all relative and absolute transformations.}\label{Fig1}
\kern-0.85em
\end{figure}

\section{Related Work}
Motivated by the success in computer vision, many deep learning based methods have now been applied to the video prediction task \cite{VideoPredictionReview}. Early approaches attempted to directly predict the pixel values of future frames \cite{Srivastava2015}. Mathieu \textit{et al.} \cite{Mathieu2016} extended upon this by training a video prediction model in adversarial fashion. Explicitly modelling the transformations between images has also been explored by Memisevic \textit{et al.} \cite{Memisevic2010, Memisevic2011}, who extract transformations between image pairs using bilinear models with multiplicative interactions. Michalski \textit{et al.} \cite{Michalski2014} introduced the Predictive Gated Pyramid (PGP), where gated autoencoders are used to model transformations between frames. They extend this idea by stacking gating units to form a recurrent pyramid architecture that inspired the design studied in this paper. The Fourier transform (FT) has long been employed as a technique for image registration \cite{Reddy1996}, but has only recently been integrated into frameworks for video prediction \cite{Farazi2019, Wolter2020}, resulting in interpretable models with few parameters. This technique has been applied to predict linear motion of pre-segmented objects and for motion segmentation.

In contrast, our approach allows for the prediction of non-linear motion and makes use of object relations. Hence, we address two problems by first predicting the motion of the individual objects and second deriving the explicit relationships between the observed objects. To this end, we extract higher-order transformations and model their evolution with a simple motion model. Results on synthetic datasets show that our model can successfully predict future frames and simultaneously extract a graph of object relations. The proposed model compares favorably to a regular gated recurrent unit (GRU) on the investigated tasks.

\section{Model}
Our model learns to produce future video frames of interacting objects by simultaneously modeling the transformations between frames and the relations between objects in a scene. The main architecture is shown in Figure \ref{Fig1}.
\subsection{Frequency-Domain Motion Estimation}
The Fourier transform has the property that a translation in the spatial domain is equivalent to a phase difference in the frequency domain \cite{ImageRegistrationReview}. Therefore, the translation of an object in consecutive frames can be represented as the phase difference of the frequency-domain representation of both frames, which can be obtained via the normalized cross-correlation:
\begin{eqnarray}
V^{(t-1:t)} = \frac{X^{(t-1)} \odot \overline{X^{(t)}}}{||X^{(t-1)} \odot \overline{X^{(t)}}||},
\label{Eq.1}
\end{eqnarray}
where \(X^{(t)} = \mathcal{F}(x^{(t)})\) is the two-dimensional discrete Fourier transform of image \(x^{(t)}\) and \(\odot\) is the Hadamard product. \(V\) can be interpreted as the momentary velocity that takes an object from its position at time \(t-1\) to the next position at \(t\). We can infer explicit representations of transformations by computing the mean difference of adjacent elements in \textit{x}- and \textit{y}-direction weighted by the energy of each frequency \cite{Farazi2021}:
\begin{eqnarray}
    \begin{pmatrix}
        v_x\\[\jot]
        v_y
    \end{pmatrix}=\frac{N}{2\pi}
    \begin{pmatrix}
        \atantwo(\Im(m_x), \Re(m_x))\\[\jot]
        \atantwo(\Im(m_y), \Re(m_y))
    \end{pmatrix}.
\label{Eq.2}
\end{eqnarray}
\paragraph{Higher-order Transformations:} If we assume constancy of the observed translation between consecutive frames, we can perpetuate this motion by repeatedly applying the extracted transformation, resulting in a linear trajectory. To model complex trajectories, we extract higher-order derivatives as illustrated in Fig. \ref{Fig1}. Keeping the highest-order transformation constant restricts the predictions to the corresponding class of polynomial trajectories, i.e. constant acceleration yields parabolic trajectories.
\paragraph{Relational Transformations:} We model binary relations, i.e., relations that hold between pairs of objects. Our goal is to study hierarchical relationships between objects in a scene, as they are ubiquitous in complex systems. These relations can be described in a directed acyclic graph (DAG) and are of the type object A \textit{"is parent of"} object B. Each relation in the adjacency matrix of the DAG is parameterized by a real number between zero and one, indicating its estimated presence in the scene. Since all transformations \(T_{o}^{(t)}\) are extracted object-wise, we can normalize the motion of each object with respect to any other object by applying the other object's inverse transformation. This results in the set of normalized transformations \(R^{(t)}\). We make all predictions in normalized space, as this allows us to reduce the potentially highly complex global motion of individual objects to simple primitive motions \emph{relative} to their parents. Given an estimate of the underlying object relation graph \(\hat{G}^{(t)}\), the relative predictions \(\hat{R}^{(t+1)}\) can be converted to global predictions \(\hat{T}_{o}^{(t+1)}\) for each object \(o\), by applying the predicted motion of the estimated parent to the object itself. This is repeated until the maximum depth of the graph is reached. The estimate of the object graph is formed during the input sequence. To this end, we evaluate the similarity of the predicted and observed transformations via the cosine-similarity of their vector representations (Eq. \ref{Eq.2}). Only predictions made for correct relationships decompose into the allowed primitive patterns of motion.\footnote{To illustrate why the similarity of predicted and observed transformations determines the object relation graph, assume we make the correct prediction that a planet moves relative to the star. The resulting predicted transformation will be highly consistent with the next observed one. This is not the case with incorrectly inferred relationships.}

\subsection{Extension by Motion Model}
We extract transformations up to degree two, resulting in the assumption of parabolic trajectories. To allow the model to predict other motion patterns, we use a residual formulation in which the motion model proposes changes to the highest order transformation. For a simple neural network to handle the obtained transformations, we calculate the average phase difference of adjacent elements in a transformation (Eq. \ref{Eq.2}), as first proposed in \cite{Farazi2021}. Instead of allowing for arbitrary motion patters, we restrict the framework to model only linear or circular paths.\footnote{\(\Delta a\) is either predicted as \(-a^{(t)}\) to force linear motion, or \(- \omega^{2}v^{(t)}\) for uniform circular motion with constant angular velocity \(\omega\).} We use small GRU to form the motion model which weights between these two modes of motion for each object. The predicted residual is \(\Delta a = c_1\Delta a_{lin} + c_2\Delta a_{cir}\), where \(c_i\) are outputs of a softmax layer. The idea of restricting the predictions to primitive motion patterns, such that the correct object trajectories can only be obtained as a combination of primitive motions and object relations, is at the core of the proposed architecture.

\section{Experimental Results}

\subsection{Datasets}
We introduce the \emph{solar system} dataset based on an underlying graph of object relations. For example, suppose an object star is parent to an object planet. Then the initial position of the star \((p_x^s, p_y^s)\) will be sampled randomly, while the planet is positioned at \((p_x^p, p_y^p) = (p_x^s, p_y^s) + (r\cos(\theta^p), r\sin(\theta^p))\) and rotates relative to its parent. To create a sample sequence, a random number of root objects are first sampled. Children are then randomly assigned to roots. This creates complex global trajectories that decompose to simple relative motions. 
We use datasets with two and three objects containing of a total of 10,000 sequences, which are randomly split into a training set (70\%), validation set (10\%), and test set (20\%).

\subsection{Evaluation}
We compare our model to a GRU trained to directly predict pixel intensities and conduct an ablation of our method by fixing the object relation graph to the identity-matrix instead of inferring it dynamically resulting in the assumption of independent objects (NoGraph). Hyper-parameters were chosen on the validation set.\footnote{The baseline GRU's hidden state size is 512 and the model is trained with a learning rate of 0.001 and batch size of 128 for 50 epochs. Our model uses a hidden state size of 64 and is trained with a learning rate of 0.01 and batch size of 32 for 1 epoch.} The source code and datasets of this work are available online.\footnote{\url{https://git.ais.uni-bonn.de/mosbach/relational_video_pred}}

\begin{figure}[ht!]
\centering
\includegraphics[width=1.0\textwidth]{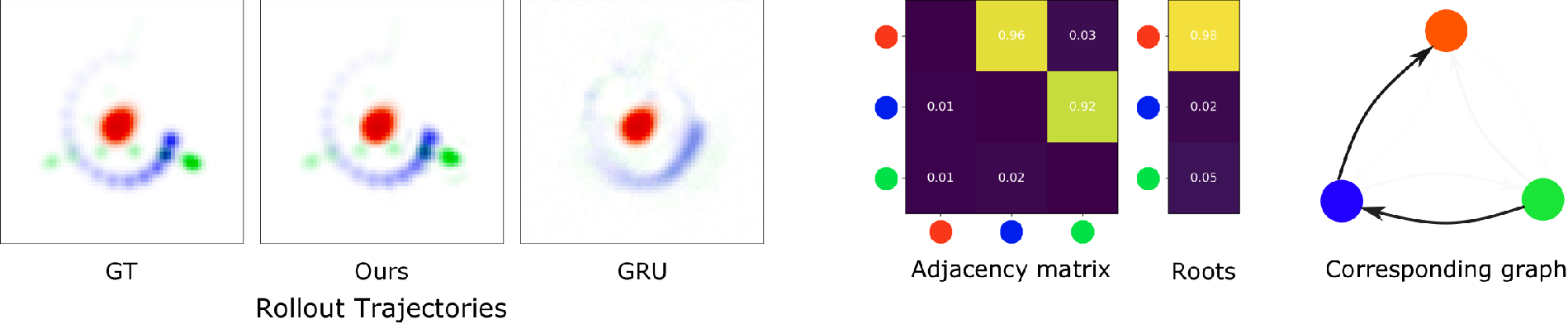}
\caption{Left: Rollout trajectories for our model and the baseline. Right: Estimated adjacency matrix and corresponding object graph. \(A_{ij}\) is equivalent to object \(i\) \emph{"is parent of"} object \(j\). Arrows indicate dependence on a parent object.}\label{Fig2}
\kern-1.5em
\end{figure}

Figure \ref{Fig2} shows sample rollouts and the inferred object relation graph which accurately represents the underlying object dependencies.

\begin{figure}[ht!]
\centering
\includegraphics[width=1.0\textwidth]{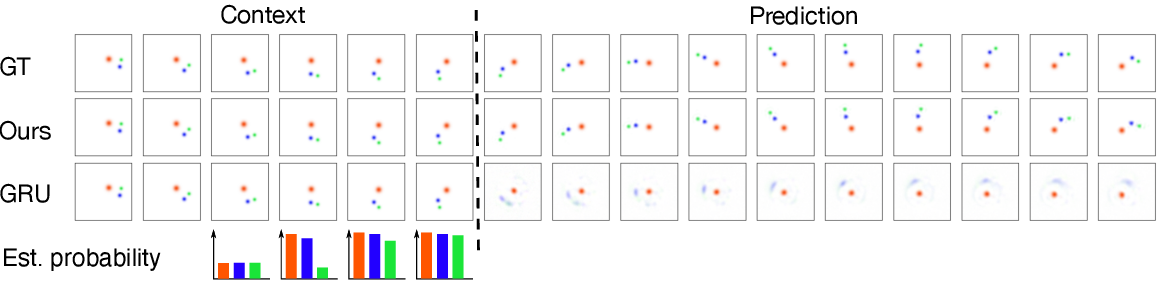}
\caption{Predictions on the solar system dataset with three objects. Below we show the estimated probability of the correct link for each object in the graph.}\label{Fig3}
\kern-0.5em
\end{figure}
In Figure \ref{Fig3} one can observe that predictions remain sharp for 10 future frames and are consistent with the ground truth (GT). The regular GRU is able to predict slow moving objects well, but considerably smears predictions for the orbiting planet, while the moon is lost entirely after the first few frames. We encourage the reader to view the videos of predictions and dynamically formed object graphs online.\(^4\)

In Table \ref{Tab.1} we compare the mean squared error (MSE) of our architecture to the baseline GRU and our model without inferrence of the relation graph. Results indicate, that our model is able to achieve improved prediction accuracy and highlight the necessity for the estimation of object relations.

\begin{table}[ht!]
  \small
  \centering
  \resizebox{\textwidth}{!}{%
    \begin{tabular}{l | c c | c c | c} \toprule
        \empty & \multicolumn{2}{c}{\textbf{2 objects}} & \multicolumn{2}{c}{\textbf{3 objects}} & \multicolumn{1}{c}{\textbf{\# parameters}} \\ \midrule
        Prediction steps & 5 & 10 & 5 & 10 & \empty \\ \midrule
        GRU & 1.5 $\pm$ 0.04 & 2.05 $\pm$ 0.09 & 5.31 $\pm$ 0.05 & 6.12 $\pm$ 0.05 & 26.0M \\
        Ours & \textbf{0.2 $\pm$ 0.04} & \textbf{1.12 $\pm$ 0.1} & \textbf{0.71 $\pm$ 0.12} & \textbf{2.79 $\pm$ 0.27} & 14.0K\\
        Ours (NoGraph) & 0.82 $\pm$ 0.01 & 3.43 $\pm$ 0.05 & 2.9 $\pm$ 0.02 & 7.26 $\pm$ 0.06 & 14.0K\\
        \bottomrule
    \end{tabular}}
  \caption{MSE prediction loss scaled by \(10^4\), averaged over at least 5 runs.}\label{Tab.1}
  \kern-0.85em
\end{table}

\section{Conclusion}
In this work, we studied a new approach to video prediction of objects related by fixed, hierarchical relationships. We were able to show how inferred frames remain sharp and are consistent with the scene dynamics. Furthermore, the formed adjacency matrices show that the model captures object dependencies, generating an interpretable representation of the observed scene. While we restrict the predicted motion to linear or circular paths here, future work could extend upon this, to allow the model to capture variable patters of motion. \vspace{1mm} \\
\emph{Acknowledgement} \begin{footnotesize}
This work was funded by grant BE 2556/16-2 of the German Research Foundation (DFG).
\end{footnotesize}

\begin{footnotesize}



\bibliographystyle{unsrt}
\bibliography{sample.bib}

\end{footnotesize}


\end{document}